\documentclass[10pt,twocolumn,letterpaper]{article}

\usepackage{wacv}              

\usepackage{algorithm}
\usepackage{algorithmic}
\usepackage{booktabs}
\usepackage{multirow}
\usepackage{rotating}

\definecolor{wacvblue}{rgb}{0.21,0.49,0.74}
\usepackage[pagebackref,breaklinks,colorlinks,allcolors=wacvblue]{hyperref}

\def\wacvPaperID{2957} 
\def\confName{WACV}
\def\confYear{2026}

\title{Learning Group Actions In Disentangled Latent Image Representations}

\author{Farhana Hossain Swarnali\textsuperscript{\dag}, Miaomiao Zhang$^{1,2 *}$, Tonmoy Hossain\textsuperscript{1*}\\
Genuity Systems Limited\textsuperscript{\dag}, Computer Science\textsuperscript{1}, Electrical and Computer Engineering\textsuperscript{2} \\
School of Engineering and Applied Science, University of Virginia\textsuperscript{*}\\
{\tt\small farhanaswarnali@gmail.com\textsuperscript{\dag}, \{mz8rr,tonmoy\}@virginia.edu\textsuperscript{*}
}
}

\begin{document}
\maketitle
\begin{abstract}

Modeling group actions on latent representations enables controllable transformations of high-dimensional image data. Prior works applying group-theoretic priors or modeling transformations typically operate in the high-dimensional data space, where group actions apply uniformly across the entire input, making it difficult to disentangle the subspace that varies under transformations. While latent-space methods offer greater flexibility, they still require manual partitioning of latent variables into equivariant and invariant subspaces, limiting the ability to robustly learn and operate group actions within the representation space. To address this, we introduce a novel end-to-end framework that for the first time learns group actions on latent image manifolds, automatically discovering transformation-relevant structures without manual intervention. Our method uses learnable binary masks with straight-through estimation to dynamically partition latent representations into transformation-sensitive and invariant components. We formulate this within a unified optimization framework that jointly learns latent disentanglement and group transformation mappings. The framework can be seamlessly integrated with any standard encoder-decoder architecture. We validate our approach on five 2D/3D image datasets, demonstrating its ability to automatically learn disentangled latent factors for group actions in diverse data, while downstream classification tasks confirm the effectiveness of the learned representations. Our code is publicly available at \href{https://github.com/farhanaswarnali/Learning-Group-Actions-In-Disentangled-Latent-Image-Representations}{\tt GitHub}.

\end{abstract}
    
\section{Introduction}
\label{sec:intro}

Understanding and modeling the symmetries underlying high-dimensional image data is fundamental to learning robust, interpretable, and controllable feature representations~\cite{hossain2025mgaug,hossain2025corld}. Group actions provide a principled mathematical framework for capturing such symmetries within a given image distribution~\cite{liu2019gift,hwang2023maganet, jin2024learning}. Recent developments have focused on incorporating group-theoretic inductive priors into deep networks, including Group-Equivariant CNNs, LieConv, and SE-Transformers~\cite{cohen2016group, finzi2020generalizing, fuchs2020se}, to build equivariant representations under various transformation groups. Other approaches aim to learn group actions directly by modeling transformations between data points, enabling compositional generalization across factor combinations~\cite{winter2022unsupervised, hwang2023maganet}. While effective, these methods primarily operate in high-dimensional data space, where applying group actions uniformly to the entire input leads to high computational cost, reduced interpretability, and susceptibility to irrelevant variations~\cite{wang2025self}. Besides, such models struggle to handle scenarios where transformations are semantically localized, where only a subset of the scene is expected to change while the rest remains fixed. For example, in an image of a cat rotating around a stationary ball, the transformation should apply only to the cat, not the background. Capturing such behavior requires learning group actions on latent representations that can disentangle and isolate transformation-relevant factors from invariant content, enabling finer control and more scalable generalization.

Learning group actions in the latent space offers a semantically meaningful alternative to modeling them directly in the data domain to capture transformation-varying representations~\cite{yang2023latent,jayakumar2024tpie,hossain2025invariant}. Building upon encoder-decoder frameworks, recent approaches have demonstrated that group actions can be learned over latent representations without requiring group-specific architectural layers or constraints, achieving diverse transformation results~\cite{jin2024learning}. Nevertheless, it primarily depends on manually subdividing the latent features into transformation-sensitive versus invariant subspaces, demanding prior domain knowledge about the expected structure of latent representations that is rarely available in practice~\cite{jin2024learning}. This approach falls apart in scenarios where transformation-relevant factors are unknown or entangled with invariant content. For instance, modeling car rotation assumes that latent variables for orientation are pre-identified, which rarely holds without supervision. This limitation motivates the need for a method that can automatically identify and disentangle transformation-sensitive components during training without relying on manual partitioning of the latent space, specifically in the latent space of image representations.

To this end, we propose a novel end-to-end framework that automatically learns a transformation-relevant latent subspace for group actions, eliminating the manual partitioning requirement that limits existing approaches. Different from existing disentangled learning methods such as $\beta$-VAE and Factor-VAE~\cite{higgins2017betavae, kim2019disentanglingfactorising} that focus on separating general factors of variation in latent space, our approach specifically targets the learning of latent components that should be equivariant versus invariant under a wide variety of group actions. Our method simultaneously learns the disentanglement of latent subspace through learnable partitioning and the corresponding group action mappings via joint optimization, enabling automatic discovery of transformation-relevant representations without architectural constraints. While our approach generalizes to a wide variety of transformation groups, we focus on the rotation group for empirical validation, demonstrating improved performance in both reconstruction and downstream classification tasks. We summarize our four primary contributions as follows:

\begin{itemize}[leftmargin=1.5em]
    \item Develop an end-to-end framework that automatically identifies transformation-sensitive and invariant latent representations through learnable partitioning.
    \item Design a joint optimization approach that simultaneously learns latent disentanglement and group action mappings.
    \item Demonstrate the generalizability of our approach across both synthetic and real-world datasets.
    \item Validate the effectiveness of our automatically learned representations through comprehensive evaluation, including downstream classification tasks.
\end{itemize}

Our proposed framework can be efficiently incorporated into any encoder-decoder architecture to learn group actions within the latent representation space. To validate our approach, we conduct experiments across five diverse 2D/3D image datasets, covering both synthetic and real-world scenarios, including high-dimensional volumetric medical images. Experimental results demonstrate that our method effectively learns transformation-specific latent partitions, achieving improved transformation control and representation quality across a broad range of
visual domains.

\section{Related Works}
\label{sec:related_works}

\noindent \textbf{Group-Equivariant Networks: }
Group-equivariant neural networks incorporate symmetry constraints directly into network architectures through specialized layers and convolution operations. These approaches leverage group representation theory to build equivariant representations under various transformation groups~\cite{cohen2016group,ruhe2023clifford, pearce2023brauer, weiler2021generale2equivariantsteerablecnns, weiler2018learning, satorras2021n, pearce2023graph}, including steerable CNNs for continuous groups~\cite{cohen2016steerable}, SE(3)-transformers for 3D transformations~\cite{fuchs2020se}, and quantum equivariant networks~\cite{Nguyen_2024}. While effective, these methods require group-specific architectural constraints and mostly operate in high-dimensional data space.


\noindent \textbf{Group Action on Data vs. Latent Space: } Traditional approaches learn group actions directly on high-dimensional data through specialized convolution operations and data augmentation strategies~\cite{romero2020attentivegroupequivariantconvolutional, NEURIPS2021_2a79ea27}. However, applying transformations uniformly to entire inputs leads to computational inefficiency and limits the ability to handle localized transformations~\cite{winter2022unsupervised, wang2025self, hwang2023maganet}. To address these limitations, recent work has shifted toward learning group actions in latent space, enabling more flexible transformation modeling~\cite{dupont2020equivariant, yang2023latent, jin2024learning}. Despite these advances, existing latent space approaches either require manual specification of transformation-sensitive versus invariant latent components or suffer from poor learned group action quality due to inefficient group action operations in the latent space, both of which limit their practical applicability, that further requires real-world domain knowledge.

\noindent \textbf{Disentangled Representation Learning: } Another line of work addresses disentangled representation learning, seeking to decompose latent representations into independent factors of variation~\cite{chen2019isolatingsourcesdisentanglementvariational, kumar2018variationalinferencedisentangledlatent, chen2016infoganinterpretablerepresentationlearning}. Central to this approach, $\beta$-VAE and its variants enforce disentanglement through regularization that balances reconstruction fidelity against latent independence~\cite{higgins2017betavae, burgess2018understandingdisentanglingbetavae, wu2023variantional, kumar2018variationalinferencedisentangledlatent, chen2019isolatingsourcesdisentanglementvariational}. Subsequent developments include Factor-VAE's treatment of total correlation~\cite{kim2019disentanglingfactorising}, causal disentanglement methods that model factor relationships~\cite{yang2023causalvaestructuredcausaldisentanglement, fan2024causal, qi2023cmvae}, multi-$\beta$ approaches for enhanced generation~\cite{uppal2025denoisingmultibetavaerepresentation, bae2023multiratevaetrainonce}, and information-theoretic frameworks for joint discrete-continuous representations~\cite{chen2016infoganinterpretablerepresentationlearning, zhang2019information}. Despite these advances, existing disentanglement methods pursue general factor independence without considering transformation-specific structure, failing to leverage the group-theoretic principles that should govern how different latent components respond to transformations.

Unlike disentangled representation learning that decomposes latent space based on general factor independence and equivariant methods that uniformly transform entire latent spaces, our approach automatically \textit{learns transformation-specific latent partitioning and jointly optimizes group action mappings}, eliminating manual specification and architectural constraints while achieving efficient domain-agnostic transformations.

\section{Our Method}
In this section, we present a novel representation learning framework that automatically learns group actions on disentangled latent representations of image transformations. We first establish the preliminaries of group action, followed by introducing our newly developed learnable partitioning mechanism that dynamically identifies transformation-sensitive and invariant latent components. Finally, we formulate the joint optimization objective that simultaneously learns latent disentanglement and group action mappings.\\

\noindent \textbf{Preliminaries.} A group $\mathcal{G}$ is an algebraic structure consisting of a set
equipped with an associative binary operation $\circ$ that contains an identity element and inverse for each element. For any elements $\{g_1, g_2, ..., g_n\} \in \mathcal{G}$, the group satisfies: (i) \textit{closure} $g_i \circ g_j \in \mathcal{G}$ for all $g_i, g_j \in \mathcal{G}$; (ii) \textit{associativity} $(g_i \circ g_j) \circ g_k = g_i \circ (g_j \circ g_k)$ for any $i,j,k$; (iii) \textit{identity} $\exists e \in \mathcal{G}$ such that $g_i \circ e = e \circ g_i = g_i$ for all $i$; and (iv) \textit{inverse} $\forall g_i \in \mathcal{G}, \exists g_i^{-1} \in \mathcal{G}$ such that $g_i \circ g_i^{-1} = g_i^{-1} \circ g_i = e$, where $e$ denotes the identity element of $\mathcal{G}$. 

\subsection{Learning Group Action on Disentangled Latent Space}

\noindent \textbf{Problem Formulation.} Consider an encoder-decoder framework, parameterized by $(\phi, \theta)$, denoted as $E_\phi: \mathcal{X} \rightarrow \mathcal{Z}$, that is responsible for mapping images $x \in \mathcal{X}$ to latent representations $z \in \mathcal{Z} \subset \mathbb{R}^d$, and decoder $D_\theta: \mathcal{Z} \rightarrow \mathcal{X}$ for reconstruction. Given a group $\mathcal{G}$ acting on image space via transformation $T_g: \mathcal{X} \rightarrow \mathcal{X}$ for $g \in \mathcal{G}$, the fundamental challenge is to learn a corresponding latent transformation $\Phi_g: \mathcal{Z} \rightarrow \mathcal{Z}$ such that
\begin{equation}
D_\theta(\Phi_g(E_\phi(x))) \approx T_g(x)
\end{equation}

Unlike existing approaches that either apply $\Phi_g$ to the entire latent space or require manual specification of transformation-sensitive dimensions, we seek to automatically learn an optimal partition $\mathcal{Z} = \mathcal{Z}_v \oplus \mathcal{Z}_i$, where $\mathcal{Z}_v$ contains transformation-varying factors and $\mathcal{Z}_i$ contains invariant factors. The objective is to learn $\Phi_g$ such that it acts selectively on $\Phi_g(z) = [\Phi_g^v(z_v); z_i]$, where $z_v \in \mathcal{Z}_v$ undergoes transformation while $z_i \in \mathcal{Z}_i$ remains invariant, enabling localized semantic transformations without affecting the invariant content.\\

\noindent \textbf{Adaptive Latent Disentanglement (ALD) Module.} Existing approaches require manual specification of which latent dimensions should be transformation-sensitive versus invariant, fundamentally limiting their applicability due to the requirement for domain expertise~\cite{jin2024learning}. We address this limitation by introducing a learnable module that automatically disentangles the latent space into transformation-relevant and invariant subspaces.

Given latent representation $z$ derived from encoder $E_\phi(x)$ taking input $x$, our disentanglement module learns to automatically partition the latent space. We parameterize this through learnable logits $\alpha \in \mathbb{R}^d$ and generate channel-wise binary assignments as

\begin{equation}
M = \mathbb{I}(\sigma(\alpha) > \tau),
\label{tau_indicator}
\end{equation}
where $\sigma$ is the sigmoid function, $\tau$ is a threshold parameter, and $\mathbb{I}$ is the indicator function. To maintain differentiability during backpropagation, we employ the straight-through estimator, yielding
\begin{equation}
z_v = M \odot z, \quad z_i = (1-M) \odot z,
\end{equation}
where $z_v$ and $z_i$ contain variant and invariant factors, respectively, with respect to the group action $g$, and $\odot$ denotes element-wise Hadamard multiplication. This module enables automatic learning of semantically meaningful latent disentanglement without architectural constraints or manual intervention. Unlike existing disentanglement methods that rely on statistical independence assumptions, our module enforces strict functional separation based on the operated group action, ensuring that it affects only relevant semantic factors while preserving invariant content.\\

\noindent \textbf{Operating Group Actions.} With our learned partition $\mathcal{Z} = \mathcal{Z}_v \oplus \mathcal{Z}_i$, we define selective group transformations that operate exclusively on transformation variant factors while preserving invariant content. For any group element $g \in \mathcal{G}$, we formulate
\begin{equation}
\Phi_g(z) = [z_v^g; z_i],
\end{equation}
where $z_v^g = \Phi_g^v(z_v)$ represents the transformed variant factors and $z_i$ remains unchanged by the transformation. The transformation function $\Phi_g^v: \mathcal{Z}_v \rightarrow \mathcal{Z}_v$ can be implemented as neural networks for complex non-linear transformations or as direct geometric operations for simple group actions like rotations. Importantly, our approach preserves fundamental group properties, including composition $\Phi_{g_1 \circ g_2}^v(z_v) = \Phi_{g_1}^v(\Phi_{g_2}^v(z_v))$, identity $\Phi_e^v(z_v) = z_v$, and inverse $\Phi_{g^{-1}}^v(\Phi_g^v(z_v)) = z_v$. This selective application enables semantically meaningful transformations on automatically discovered factors while maintaining the integrity of invariant features. \\

\begin{figure*}
    \centering
    \includegraphics[width=0.9\textwidth]{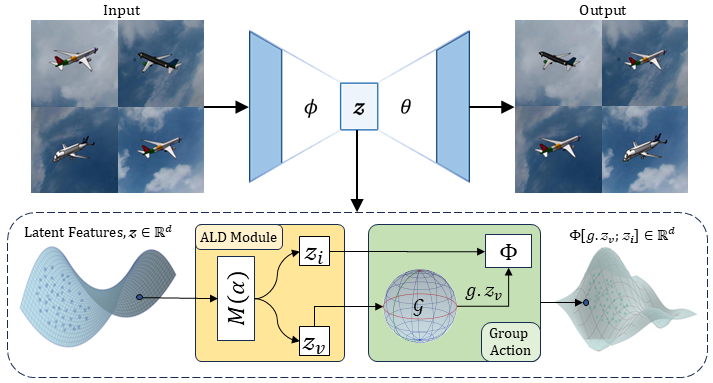}
    \caption{An overview of our proposed network to learn group actions in disentangled latent image representations.}
    \label{fig:model}
\end{figure*}

\subsection{Network Loss and Joint Optimization} 
Our framework jointly optimizes the encoder-decoder and disentanglement module parameters $(\phi, \theta, \alpha)$ through a unified training objective. Since we seek to learn transformations that behave consistently across image and latent spaces while maintaining meaningful separation between variant and invariant factors, we formulate our network loss as

\begin{equation}
\mathcal{L}_{total} = \lambda_r \mathcal{L}_{recon} + \lambda_i \mathcal{L}_{inv} + \lambda_v \mathcal{L}_{const},
\end{equation}
where the reconstruction loss $\mathcal{L}_{recon} = \|D_\theta(\Phi_g(E_\phi(x))) - T_g(x)\|^2$ ensures our latent transformations produce equivalent results to image-space transformations. To explicitly enforce that invariant factors truly remain unchanged, we introduce the invariant loss as $\mathcal{L}_{inv} = \|z_i^{(x)} - \text{sg}[z_i^{(T_g(x))}]\|^2$, which constrains $z_i$ extracted from both original and transformed images to be identical. The consistency loss $\mathcal{L}_{const} = \|\Phi_g^v(z_v^{(x)}) - \text{sg}[z_v^{(T_g(x))}]\|^2$ ensures that transforming variant factors in latent space matches the variant factors from actually transformed images. Here, $\text{sg}[\cdot]$ prevents gradient flow through the encoder, thereby focusing the learning on latent space transformations rather than feature extraction modifications. We set the hyperparameters $\lambda_r, \lambda_i$, and $\lambda_v$ to balance reconstruction quality with disentanglement strength, while the threshold $\tau$ (Eq.~\ref{tau_indicator}) controls the sparsity of the learned partition. We jointly optimize all the network parameters, including the ALD and group action learning modules, enabling the automatic discovery of meaningful latent partitions while simultaneously learning their corresponding group actions. We summarize the optimization steps for jointly learning the disentanglement module and group transformations in Alg.~\ref{alg:group_action_learning}.

\begin{algorithm}[h]
\caption{Learning Group Actions on Disentangled Latent Space}
\label{alg:group_action_learning}
\begin{algorithmic}[1]
\REQUIRE Dataset $\mathcal{D} = \{(x_i, T_{g_i}(x_i), g_i)\}_{i=1}^N$, hyperparameters $\lambda_r,\lambda_i, \lambda_v, \tau$
\ENSURE Trained encoder $E_\phi$, decoder $D_\theta$, disentanglement logits $\alpha$
\STATE Initialize parameters $\phi, \theta, \alpha$
\FOR{epoch = 1 to max\_epochs}
    \FOR{$(x, T_g(x), g)$ in $\mathcal{D}$}
        \STATE $z \leftarrow E_\phi(x)$
        \STATE $z^{T_g} \leftarrow \text{sg}[E_\phi(T_g(x))]$ \COMMENT{Stop gradient for target}
        \STATE $M \leftarrow \mathbb{I}(\sigma(\alpha) > \tau)$
        \STATE $z_v \leftarrow M \odot z$, $z_i \leftarrow (1-M) \odot z$
        \STATE $z_v^{T_g} \leftarrow M \odot z^{T_g}$, $z_i^{T_g} \leftarrow (1-M) \odot z^{T_g}$
        \STATE $z_v^g \leftarrow \Phi_g^v(z_v)$
        \STATE $\mathcal{L}_{recon} \leftarrow \|D_\theta([z_v^g; z_i]) - T_g(x)\|^2$
        \STATE $\mathcal{L}_{inv} \leftarrow \|z_i - z_i^{T_g}\|^2$
        \STATE $\mathcal{L}_{const} \leftarrow \|z_v^g - z_v^{T_g}\|^2$
        \STATE $\mathcal{L}_{total} \leftarrow 
        \lambda_r \mathcal{L}_{recon} + \lambda_i \mathcal{L}_{inv} + \lambda_v \mathcal{L}_{const}$
        \STATE Update $\phi, \theta, \alpha$ using $\nabla \mathcal{L}_{total}$
    \ENDFOR
\ENDFOR
\RETURN $E_\phi, D_\theta, \alpha$
\end{algorithmic}
\end{algorithm}

\section{Experimental Results}

\subsection{Datasets} 
To validate the development of our group action learning model on disentangled latent image representations and assess its efficiency, we conduct extensive experiments on diverse 2D and 3D synthetic and real computer vision datasets, including 2D synthetic digits, real 3D brain MRIs, 3D adrenal shapes, and 3D objects with diverse backgrounds.

\noindent \textbf{2D Rotated MNIST.} We use a variant of the original 2D MNIST dataset~\cite{lecun1998mnist}, obtained by randomly rotating each image with angles $\theta \sim \mathcal{U}[0, 2\pi)$. The grayscale dataset consists of $70000$ handwritten digits across $10$ classes $(0-9)$, with each image of size $28 \times 28$.

\noindent \textbf{2D Rotated and Blocked MNIST.} We further modify the synthetic Rotated MNIST dataset by introducing a random probabilistic white square block of size $7^2$ on the images to analyze the performance of our framework under occlusion settings~\cite{jin2024learning}.

\noindent \textbf{3D Brain MRIs.} We use the OASIS-I Brain MRI dataset \cite{Hoopes_2022,marcus2007open}, consisting of $414$ T1-weighted MRI scans. All MRIs are preprocessed to $160 \times 192 \times 224$ with $1.25\,\text{mm}^3$ isotropic voxels, including skull-stripping, intensity normalization, bias field correction, and affine pre-alignment.

\noindent \textbf{3D Adrenals Shapes.} We utilize AdrenalMNIST3D \cite{yang2023medmnist} dataset that consists of $1584$ 3D shapes of left and right adrenal glands, each of size $64^3$ taken from $792$ patients. The images contain either normal adrenal glands or adrenal mass lesions.

\noindent \textbf{3D Objects with Diverse Backgrounds.} We select airplane objects from ShapeNet Core~\cite{chang2015shapenet} against sky backgrounds from SWIMSEG~\cite{dev2016color}. Following Jin et al.~\cite{jin2024learning}, we apply $100$ uniformly sampled rotations from SO(3) to generate image sets. We maintain consistent backgrounds for fair comparison while analyzing robustness across diverse sky backgrounds.

\subsection{Experiments}

\noindent \textbf{Evaluation of Learned Group Action.} We first assess the performance of learning group actions in the disentangled latent representation by evaluating the predicted images transformed by specific group action. For the group actions $g\in\mathcal{G}$, we utilize either SO(2) or SO(3) rotation action groups, depending on the dataset. We extract the varying features $z_v$ from the latent representation $z$ using our adaptive latent disentanglement module and then apply $g$ on it, where we can define the group action operation as $g \cdot z = [g \cdot z_v; z_i]$. To evaluate the reconstruction quality, we measure PSNR, SSIM, and RMSE on randomly sampled rotation angles $\theta$ and further qualitatively compare the reconstructed images.

\noindent \textbf{Ablation Study on Network Loss.} To evaluate the impact of each component in enabling our model to learn group actions in latent space, we perform an ablation study by systematically removing the invariant loss $\mathcal{L}_{inv}$ and consistency loss $\mathcal{L}_{const}$ components. We evaluate each configuration by quantitatively reporting reconstruction quality metrics across both 2D/3D datasets, and also qualitatively evaluate the reconstructed images across all settings.

\noindent \textbf{Effectiveness of ALD Module on Learning Group Actions.} To further validate that our model successfully disentangles variant features from invariant ones with respect to group actions, we perform downstream classification tasks using different components of the learned latent space. We compare classification performance using the full latent representation $z$, variant features $z_v$, and invariant features $z_i$. Successful disentanglement is indicated when $z_v$ achieves similar performance to $z$, while $z_i$ shows degraded performance, demonstrating that group-specific transformations are captured in the variant component.

We further examine the sensitivity of the threshold parameter $\tau$, a key design component of our ALD module. Additionally, we demonstrate the effectiveness of our disentanglement module through feature swapping experiments, where we interchange the $z_v$ and $z_i$ components between different input latent representations and analyze the decoded outputs to validate the separation of variant and invariant information.

\subsection{Baseline Selection and Implementation Details}

We compare our approach against five state-of-the-art models: (i) EquivNR~\cite{dupont2020equivariant} enforces 3D transformation equivariance in scene representations; (ii) InfoGAN-CR~\cite{lin2020infogan} learns disentanglement through contrastive regularization with random latent sampling; (iii) CausalVAE~\cite{fan2024causal} employs causal mechanisms to disentangle independent factors; (iv) MAGANet~\cite{hwang2023maganet} learns disentangled representations in a variational framework; and (v) GALR~\cite{jin2024learning} learns group transformations by manually partitioning the latent space. Notably, except for EquivNR and GALR, the other methods primarily focus on disentanglement learning rather than explicitly learning group actions. Our key contribution lies in automatically disentangling latent representations to isolate group-specific transformations, making us \textit{the first to learn group actions within a disentangled latent space}.

Our architecture employs an encoder-decoder framework with convolutional downsampling modules in the encoder and corresponding upsampling modules in the decoder. During training, we randomly sample pairs of data points from the training distribution and optimize the model using MSE reconstruction loss with weighting factors $\lambda_r$, $\lambda_i$, and $\lambda_v$ set to $1$. All models are trained using the Adam optimizer.

\subsection{Results}

Tab.~\ref{tab:comparison} demonstrates that our method achieves superior performance compared to existing approaches across both 2D and 3D datasets. Our model consistently outperforms all baselines across PSNR, SSIM, and RMSE metrics, showcasing the effectiveness of our adaptive disentanglement mechanism. While EquivNR enforces transformation equivariance and GALR manually partitions latent space, InfoGAN-CR and CausalVAE focus primarily on general disentanglement, resulting in lower performance due to their lack of transformation-specific structure. Notably, compared to GALR's manual partitioning approach, our automatic discovery mechanism achieves better performance across all evaluated metrics. The substantial performance gains across all datasets validate our learnable partitioning mechanism as an effective solution for group action learning in the latent spaces.


\begin{table}[h]
\caption{Performance comparison of learning group actions across all models on 2D and 3D datasets. (RMNIST: 
Rotated MNIST; RBMN: Rotated and Blocked MNIST)}
\resizebox{\linewidth}{!}{\begin{tabular}{lccccccc}
\toprule
& \textit{Metrics}     & \textbf{EquivNR} & \textbf{InfoGAN-CR} & \textbf{CausalVAE} & \textbf{MAGANet} & \textbf{GALR} & \textbf{Ours} \\ \midrule
\multirow{3}{*}{\rotatebox{90}{RMNIST}}   & PSNR & $21.77$       & $17.39$     & $24.02$   &    $18.63$     & $18.63$ & $\mathbf{30.61}$ \\ \cmidrule{3-8}
& SSIM & $0.87$       & $0.65$     & $0.92$    &    $0.63$     & $0.65$  & $\mathbf{0.98}$ \\ \cmidrule{3-8}
& RMSE & $0.16$       & $0.29$     & $0.13$    &   $0.31$      & $0.28$  & $\mathbf{0.06}$ \\ \midrule
\multirow{3}{*}{\rotatebox{90}{RBMN}}     & PSNR & $17.30$       & $15.08$    & $21.02$   &    $17.53$     & $18.62$ & $\mathbf{27.15}$ \\ \cmidrule{3-8}
& SSIM & $0.71$       & $0.53$     & $0.84$     &    $0.57$     & $0.66$  & $\mathbf{0.95}$ \\ \cmidrule{3-8}
& RMSE & $0.33$       & $0.37$     & $0.19$    &    $0.34$     & $0.28$  & $\mathbf{0.10}$ \\ \midrule
\multirow{3}{*}{\rotatebox{90}{Adrenals}} & PSNR & $21.77$        & $21.90$    & $23.08$   &   $19.83$      & $23.09$ & $\mathbf{28.91}$ \\ \cmidrule{3-8}
& SSIM & $0.87$         & $0.88$     & $0.91$    &  $0.81$       & $0.88$  & $\mathbf{0.97}$ \\ \cmidrule{3-8}
& RMSE & $0.16$         & $0.17$     & $0.14$    &    $0.16$     & $0.15$  & $\mathbf{0.07}$ \\ \bottomrule
\end{tabular}}
\label{tab:comparison}
\end{table}

\begin{table}[htbp]
\caption{Ablation study of our model based on different loss terms.}
\centering
\resizebox{\linewidth}{!}{\begin{tabular}{lccccc}
\toprule
& $\mathcal{L}_{inv}$ & $\mathcal{L}_{const}$ & \textit{PSNR} $(\uparrow)$ & \textit{SSIM} $(\uparrow)$ & \textit{RMSE} $(\downarrow)$ \\
\midrule
\multirow{3}{*}{\rotatebox{90}{RBMN}} & \checkmark       & $\times$          &   $\mathbf{27.16}$   &  $0.95$    &  $\mathbf{0.097}$    \\
&  $\times$      & \checkmark         &  $27.12$    &   $0.95$   &  $0.098$    \\
&  \checkmark      &   \checkmark       &   $27.15$   &  $0.95$    &  $\mathbf{0.097}$    \\
\midrule
\multirow{3}{*}{\rotatebox{90}{Brains}}             & \checkmark       & $\times$        & $27.86$     &  $0.92$     &   $0.081$   \\
&  $\times$      & \checkmark        &   $27.66$   & $0.89$     &   $0.083$   \\
& \checkmark       & \checkmark         &  $\mathbf{31.64}$    &  $\mathbf{0.93}$    &  $\mathbf{0.052}$    \\
\midrule
\multirow{3}{*}{\rotatebox{90}{Adrenals}}           &   \checkmark       & $\times$        &  $28.16$   &  $0.94$    &  $0.079$    \\
&  $\times$      & \checkmark         &  $27.35$    &  $0.93$    &   $0.087$   \\
& \checkmark       & \checkmark         &  $\mathbf{28.91}$     &  $\mathbf{0.97}$    & $\mathbf{0.073}$    \\
\bottomrule
\end{tabular}}
\label{ablation_loss}
\end{table}

\begin{figure*}
    \centering
    \includegraphics[width=\textwidth]{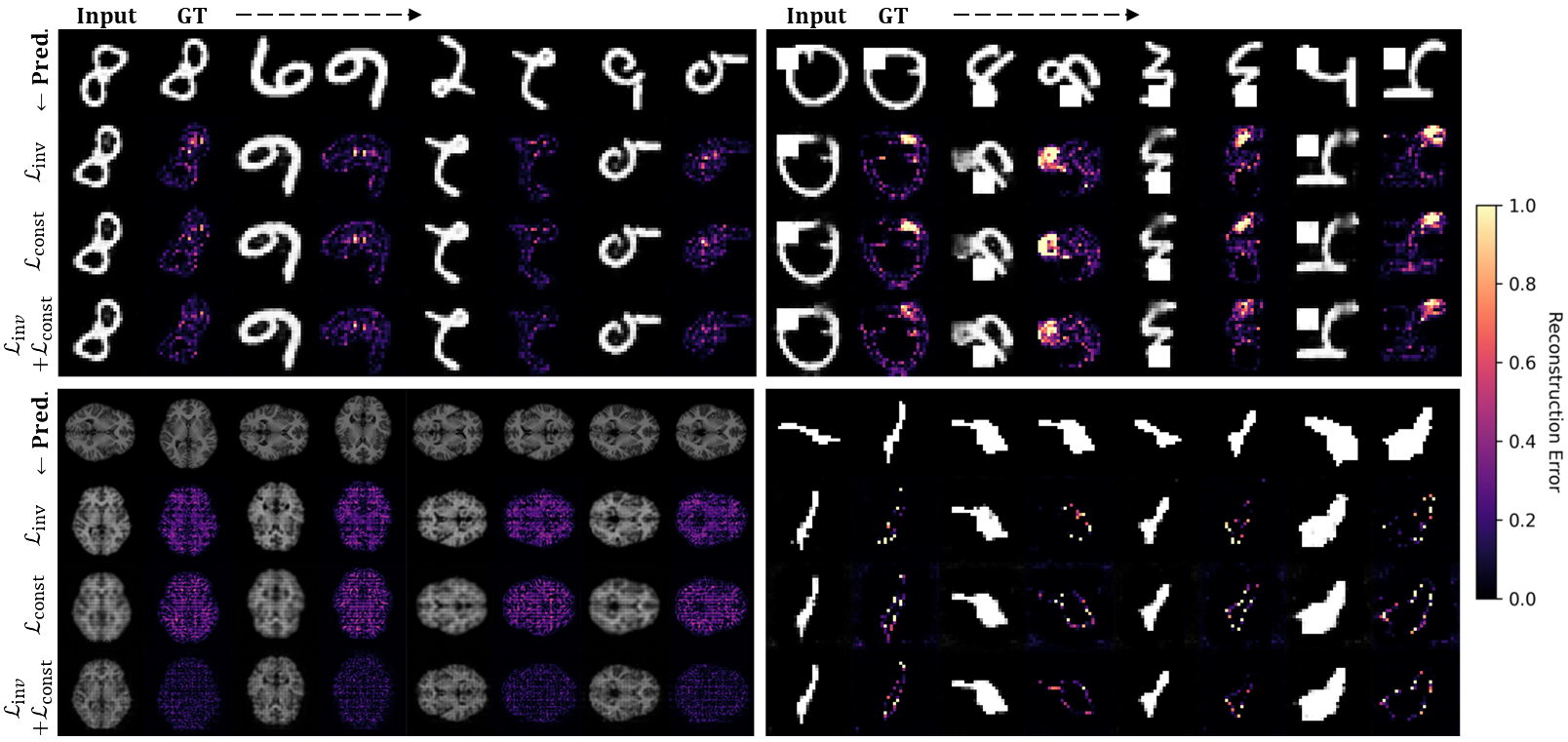}
    \caption{Qualitative ablation study of different loss components on learning group action through evaluating reconstruction quality. In each dataset panel, Input and Ground-Truth (GT) pair visualizations (from left to right) with the predicted images operated after applying the learned group actions (top to bottom). Top row: 2D rotated MNIST digit, and rotated and blocked MNIST results, learning SO(2) group actions. Bottom row: 3D brain MRIs and 3D adrenal shapes results, learning SO(3) group actions. The colormap indicates reconstruction error magnitude, with the range of $0-1$. Our model efficiently learns the underlying SO(2) and SO(3) group actions across all loss configurations, with the joint optimization ($\mathcal{L}_{inv} + \mathcal{L}_{const}$) achieving the lowest error magnitudes and most optimal reconstructions.}
    \label{fig:ablation}
\end{figure*}

Tab.~\ref{ablation_loss} presents an ablation study examining the contribution of individual loss components in our joint optimization framework. The results demonstrate the effectiveness of combining both invariant loss $\mathcal{L}_{inv}$ and consistency loss $\mathcal{L}_{const}$ for robust latent group action learning. Our joint optimization framework consistently achieves optimal or near-optimal performance across all datasets, with particularly notable improvements on complex 3D medical imaging datasets. For the Brain experiments, our combined approach yields substantial improvements in reconstruction quality, demonstrating the positive effect of enforcing both invariant factor stability and transformation consistency simultaneously. These results validate our network loss formulation, showing that the joint optimization of invariance and consistency constraints provides robust performance across diverse data modalities and transformation complexities.

Fig.~\ref{fig:ablation} provides qualitative visualization of learned group actions across different loss configurations, where the colormap indicates reconstruction error magnitude. The results demonstrate effective learning of SO(2) rotations on MNIST experiments and SO(3) rotations on brain and adrenal experiments across all configurations. Individual loss terms ($\mathcal{L}_{inv}$ or $\mathcal{L}_{const}$ alone) produce noticeable artifacts and higher error magnitudes, while optimizing the combined loss terms ($\mathcal{L}_{inv} + \mathcal{L}_{const}$) achieves significantly cleaner reconstructions with lower-magnitude errors.

\begin{figure*}
    \centering
    \includegraphics[width=\linewidth]{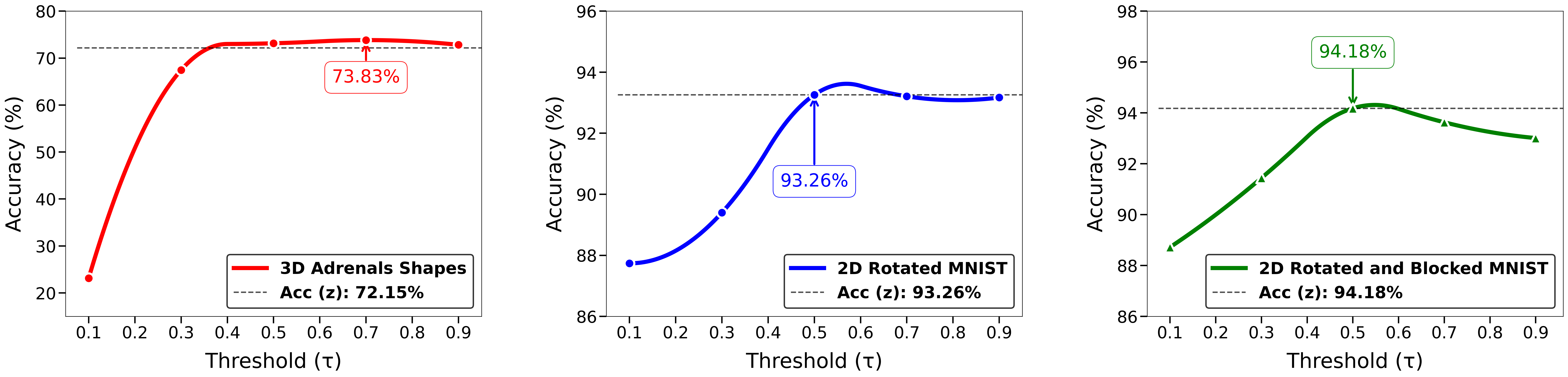}
    \caption{Classification performance using varying representation $z_v$ across different mask threshold values $\tau$ for automatic latent disentanglement on 3D Adrenal and 2D MNIST datasets (from left to right). The threshold $\tau$ determines the binary partition between transformation-varying and invariant latent factors through $M = \mathbb{I}(\sigma(\alpha) > \tau)$. Optimal performance around $\tau = 0.5-0.6$ validates automatic discovery of meaningful disentangled representations.}
    \label{fig:threshold_acc}
\end{figure*}

While brain MRI reconstruction intensity appears limited due to our basic encoder-decoder architecture, our primary objective is learning group actions in disentangled latent space rather than improving reconstruction fidelity. The error visualizations clearly show that our joint optimization produces more consistent transformation quality across all samples, validating our quantitative findings and demonstrating effective latent group action learning.

\begin{table}[htbp]
\caption{Effectiveness of learned feature representation in the latent space of image representations.}
\centering
\resizebox{\linewidth}{!}{%
\begin{tabular}{lcccccccc}
\toprule
& $z$ & $z_i$ & $z_v$ & \textit{Acc} & \textit{Prec} & \textit{Rec} & \textit{F1-sc} & \textit{AUC} \\ \midrule
\multirow{3}{*}{\rotatebox{90}{RMNIST  }}    & \checkmark  & $\times$     & $\times$     &  $92.68$   & $92.93$     &  $92.68$   & $92.68$      &  $99.61$   \\ \cmidrule{2-9}
&   $\times$  &   \checkmark   &   $\times$   & $19.75$    &  $17.83$    &  $19.75$   &  $10.48$     &  $82.37$   \\ \cmidrule{2-9}
& $\times$  &  $\times$    &  \checkmark     &  $\mathbf{93.26}$   &  $\mathbf{93.47}$    & $\mathbf{93.26}$    &  $\mathbf{93.28}$     &  $\mathbf{99.66}$ \\ \midrule
\multirow{3}{*}{\rotatebox{90}{RBMN   }}     & \checkmark  &  $\times$    &   $\times$   &  $93.27$   &   $93.46$   & $93.27$   &  $93.23$     &  $99.72$   \\ \cmidrule{2-9}
& $\times$  &   \checkmark  &    $\times$   &  $44.18$   &   $48.26$   &  $44.18$   & $40.09$      &  $84.79$  \\ \cmidrule{2-9}
&  $\times$ &   $\times$    &   \checkmark  &  $\mathbf{94.18}$   &   $\mathbf{94.32}$   &  $\mathbf{94.18}$   &  $\mathbf{94.18}$ & $\mathbf{99.75}$    \\ \midrule
\multirow{3}{*}{\rotatebox{90}{Adrenals   }} & \checkmark  & $\times$     &  $\times$    &  $72.15$   &  $81.73$    &  $72.15$   &  $74.26$     &  $81.05$  \\ \cmidrule{2-9}
&  $\times$  &  \checkmark    &   $\times$   &  $33.22$   &  $80.36$    & $33.22$    &   $27.65$    & $76.46$  \\ \cmidrule{2-9}
&  $\times$  &   $\times$  &     \checkmark  &  $\mathbf{73.15}$   &  $\mathbf{84.18}$    &  $\mathbf{73.15}$   &  $\mathbf{75.24}$     & $\mathbf{84.20}$ \\ \bottomrule  
\end{tabular}}
\label{clf_results}
\end{table}

Tab.~\ref{clf_results} evaluates the effectiveness of our learned disentangled latent representations through classification performance using different latent components. The results demonstrate that our adaptive latent disentanglement module efficiently partitions the latent space into meaningful transformation-varying ($z_v$) and invariant ($z_i$) subspaces. Notably, classification using only the varying features $z_v$ achieves comparable or slightly better performance than the complete latent representation $z$ across all datasets, indicating that most discriminative information is concentrated in the varying subspace. This suggests our disentanglement mechanism effectively captures task-relevant factors while removing redundant components. Conversely, classification performance using only invariant features $z_i$ drops dramatically across all datasets, confirming that our method optimally isolates transformation-invariant information with limited discriminative capacity. These results validate that our approach effectively separates transformation-specific discriminative features from invariant factors without manual specification.

Fig.~\ref{fig:threshold_acc} visualizes the impact of threshold parameter $\tau$ on classification performance using transformation-varying features $z_v$, demonstrating automatic learning of disentangled representations. The threshold $\tau$ controls partition sparsity by determining which latent dimensions are assigned to varying ($z_v$) versus invariant ($z_i$) subspaces through $M = \mathbb{I}(\sigma(\alpha) > \tau)$. Results reveal optimal performance around $\tau = 0.5-0.6$ across 2D/3D experiments, where $z_v$ achieves comparable performance to complete representation $z$ (dashed baselines: $72.15\%$, $93.26\%$, and $94.18\%$). Lower thresholds ($\tau < 0.3$) assign excessive dimensions to the varying subspace, including irrelevant factors that degrade the performance. Higher thresholds ($\tau > 0.7$) create overly sparse representations, limiting discriminative capacity of the latent disentanglement module. These outcomes validates the impact of the threshold parameter $(\tau)$ to the performance of latent disentanglement.

\begin{figure}[htbp]
    \centering
    \includegraphics[width=0.9\linewidth]{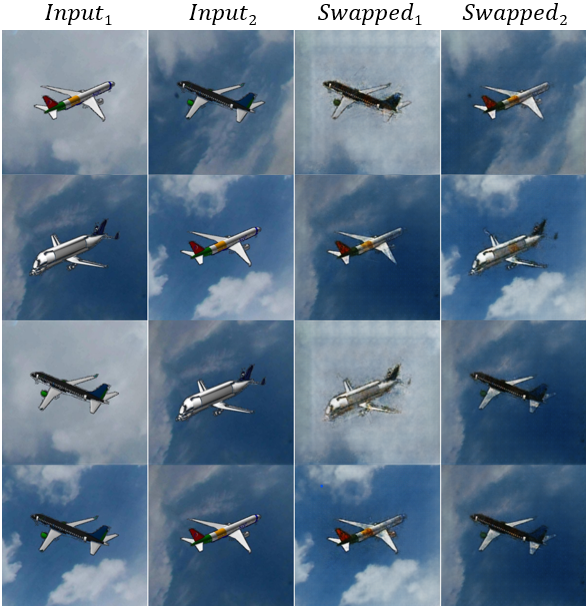}
    \caption{Visualizations of latent factor swapping for the 
    validation of learned disentanglement. By exchanging varying ($z_v$) and invariant ($z_i$) components between input pairs, our method generates novel combinations of object orientations with different backgrounds, demonstrating effective separation of transformation-sensitive and invariant factors in latent spaces.}
    \label{fig:plane_swap}
\end{figure}

Fig.~\ref{fig:plane_swap} demonstrates the effectiveness of our learned disentanglement through latent factor swapping experiments. We extract varying ($z_v$) and invariant ($z_i$) components from two airplane images with different orientations and backgrounds, then create novel combinations by swapping components: combining the varying factors from one image with invariant factors from another ($[z_{v1}; z_{i2}]$ and $[z_{v2}; z_{i1}]$). The swapped outputs generate semantically coherent combinations not present during training, where airplane orientations (varying factors) are correctly transferred while preserving background sky conditions and lighting (invariant factors). This demonstrates that our adaptive partitioning mechanism effectively separates transformation-sensitive features from invariant information without manual specification during the network learning, which further confirms that our framework learns meaningful group actions on disentangled latent representations.

\begin{figure}[htbp]
    \centering
    \includegraphics[width=\linewidth]{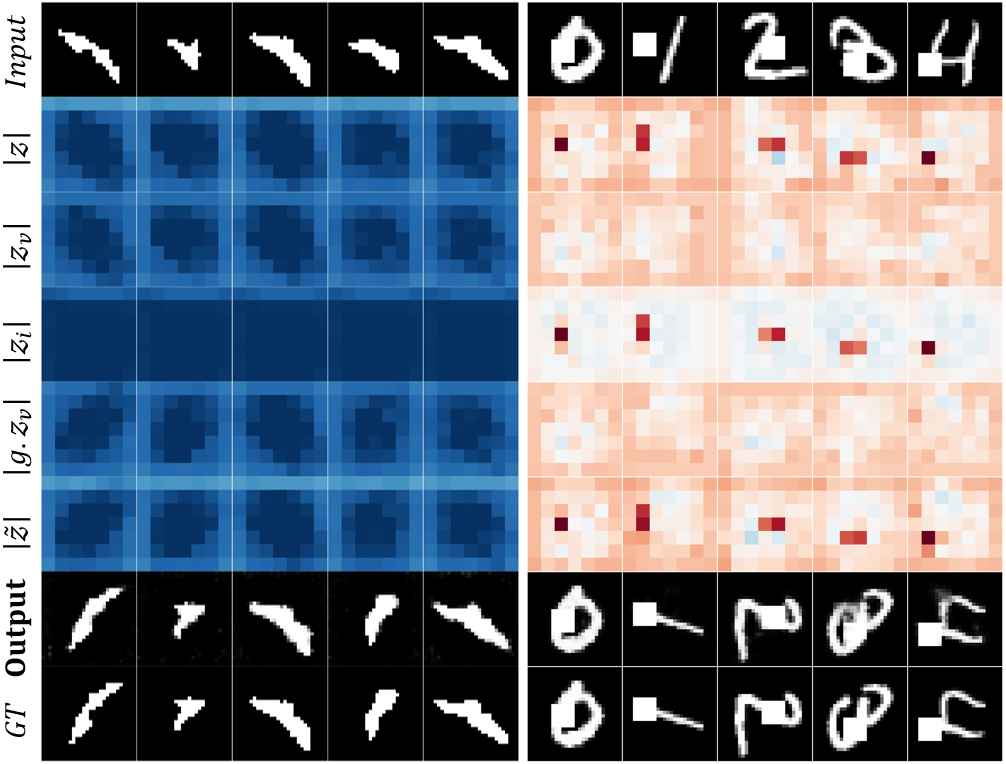}
    \caption{Visualization of disentangled latent representations showing averaged magnitude of varying ($|z_v|$) and invariant ($|z_i|$) components for 3D adrenal shapes (left) and 2D rotated and 
    blocked MNIST (right), demonstrating effective separation of transformation-specific and transformation-invariant features.}
    \label{fig:latent_vizs}
\end{figure}

Fig.~\ref{fig:latent_vizs} demonstrates effective disentanglement of transformation-sensitive and invariant features in latent space through visualizing the magnitude of $|z_v|$ and $|z_i|$ components. The varying component $|z_v|$ shows distinct activation patterns that change with different input rotations, capturing transformation-specific information. In contrast, the invariant component $|z_i|$ maintains consistent activation patterns across all input samples, confirming optimal separation of group-variant and group-invariant representations. For 3D adrenal shapes (left), $|z_v|$ activates selectively for rotation-sensitive features, while for 2D rotated 
MNIST (right), the varying component captures digit orientation changes. The stable reconstructed outputs validate that our adaptive disentanglement module captures non-essential information while effectively isolate transformation-related latent features.
\section{Conclusion}
In this paper, we present a novel framework that learns group actions on automatically disentangled latent spaces, enabling models to achieve better generalization. The automatic disentanglement of transformation-relevant latent factors from invariant factors eliminates the manual specification requirements of existing frameworks, allowing for seamless learning of group actions within the disentangled latent space. Experiments performed on diverse synthetic and real datasets validate the robustness of our framework in learning effective group actions on latent transformation-sensitive factors. The downstream classification tasks further demonstrate the framework's potential for robust applications across diverse tasks by leveraging the disentangled varying representations through the learned latent group actions. Future enhancements to our framework could address: (i) learning input image-dependent group actions for the disentanglement of latent space, (ii) applying compositions of group actions on the latent space to capture multiple robust transformations, and (iii) investigating applications of the developed framework to additional downstream tasks that benefit from group-equivariant representations.

\section*{Acknowledgments}
\noindent This work was supported by NSF CAREER Grant 2239977.

{
    \small
    \bibliographystyle{ieeenat_fullname}
    \bibliography{main}
}

\end{document}